\documentclass[letterpaper, 10pt, conference]{ieeeconf}
\IEEEoverridecommandlockouts
\usepackage{graphicx}
\usepackage{amsmath}
\usepackage{amssymb}
\usepackage{booktabs}
\usepackage{xspace}

\newcommand{\method}{\textsc{SUPER}\xspace}

\title{\LARGE\bf
SUPER -- Sensitivity-based Uncertainty-aware Performance and Risk Assessment in Visual Inertial Odometry}

\author{Johannes A.~Gaus$^{1}$, Daniel Häufle$^{1}$, and Woo-Jeong Baek$^{2}$
\thanks{$^{1}$ Hertie Institute for Clinical Brain Research \& Center for Integrative Neuroscience \& Center for Bionic Intelligence Tübingen Stuttgart (BITS), University of Tübingen, Germany.}
\thanks{$^{2}$ Artificial Intelligence Institute (AIIS), Seoul National University, Republic of Korea.}
\thanks{The authors thank the International Max Planck Research School for Intelligent Systems (IMPRS-IS) for supporting Johannes A. Gaus.}}

\begin{document}
\maketitle
\thispagestyle{empty}
\pagestyle{empty}

\begin{abstract}
This paper presents the framework SUPER (\textbf{S}ensitivity-based \textbf{U}ncertainty-aware \textbf{PE}rformance and \textbf{R}isk assessment in visual inertial odometry) that evaluates the risks of VIO controllers at runtime. While existing methods in literature present approaches to handle the occurrence of large residuals, the insensitivity to different types of visual disturbances poses a challenge in remaining the reliability of optimizers. In order to address this issue, this manuscript develops SUPER that refers to stereo observations and derives an uncertainty on the feature level by propagating the covariances per landmark. By combining the feature uncertainty with the residual, \method computes an activation signal that serves as an activation threshold for the optimizer in the subsequent time step. Practically, SUPER thereby enables to track the occurrence of undesired disturbances at runtime. To validate \method, we perform experiments across 13 stacks on different disturbation classes. By referring to sequences of the data sets EuRoC and TUM-VI, we test different hypotheses on the stability and risk-awareness of our \method. The results emphasize the real-time capability, performance and the computational efficiency of our approach.  
\end{abstract}

%VIO controllers are hard to evaluate: a nondeterministic optimizer changes trajectories from scheduling alone, so plain comparison cannot isolate a real effect. We introduce an action-verification protocol -- factor-level logs, an inactive control, and same-linearization shadow replay -- applied to \method, a bounded controller converting a track-consistency signal into per-feature information. The controller is computationally efficient ($-0.014\%$ wall time) and bounded ($0.59\%$/$7.06\%$ factors modified, clean/degraded). Across 13 evaluation splits, all confidence intervals cross zero. The held-out corruption-family decomposition is heterogeneous ($Q=20.8$, $p=0.008$), but its strongest exploratory result, motion blur, fails FDR correction ($q=0.07$ within families) and regresses prospectively ($+4.3\%$). No accuracy claim survives both correction and replication. None of four alternative controls dominates \method across median and tail behavior; a duty-matched random control shows no median benefit and a substantially heavier loss tail. \method is bounded and measurably active under degradation, but does not establish a general accuracy gain. The action-verification protocol is additionally demonstrated on a structurally different controller.

% 

%While VIO frameworks with promising performance exist in literature, approaches that are designed to react to dangerous events by adapting the behavior remain sparse. To fill this gap, we suggest \method  

%\noindent\textbf{Index Terms}---Visual--Inertial Odometry, Adaptive Estimation, Robust Estimation, Sensor Fusion.

\section{INTRODUCTION}
Modern visual odometry (VO), visual-inertial odometry (VIO), and SLAM systems achieve high accuracy on standard benchmarks~\cite{Davison2007,Mourikis2007,Leutenegger2015,Campos2021ORB-SLAM3,Teed2021}. However, visual residuals are commonly assigned to representations that are insensitive to the varying character of visual disturbances (e.g., image conditions). In fact, these residuals are represented by fixed and constant values. Practically, this may lead to a misinterpretation of the state, that in turn might result in safety-critical robot movements or even undesired collisions with the environment. While robust losses suppress large residuals, capturing feature-specific misalignment remains challenging. In case of degraded image conditions, fixed information might therefore result in a overestimation of the visual reliability. On the other hand, applying a downweighting procedure may yield to the neglection of valuable constraints under nominal conditions. In order to overcome this conflict and derive a reasonable compromise within the controller, this manuscript presents the framework SUPER (Sensitivity-based Uncertainty-aawre PErformance and Risk assessment) that derives a measure for the track consistency on the basis of uncertainties and implements it into a bounded controller. By studying whether the observed changes in the trajectory can be attributed to the input, we aim to consider the fact that the controller operates inside a nondeterministic optimizer. In fact, a naive comparison may be insufficient because enabling a controller can alter solver scheduling even in the absence of modifications on the visual factors. We therefore pair trajectory evaluation with factor-level action logs, an inactive-instrumented control, and a fixed information shadow solve from the same Ceres or g2o linearization. By introducing this action-verification protocol, we hope to distingush state increments induced by the controller from the frontend, initialization, and scheduling variability.
\begin{figure}[t]
    \centering
    \includegraphics[scale=0.38]{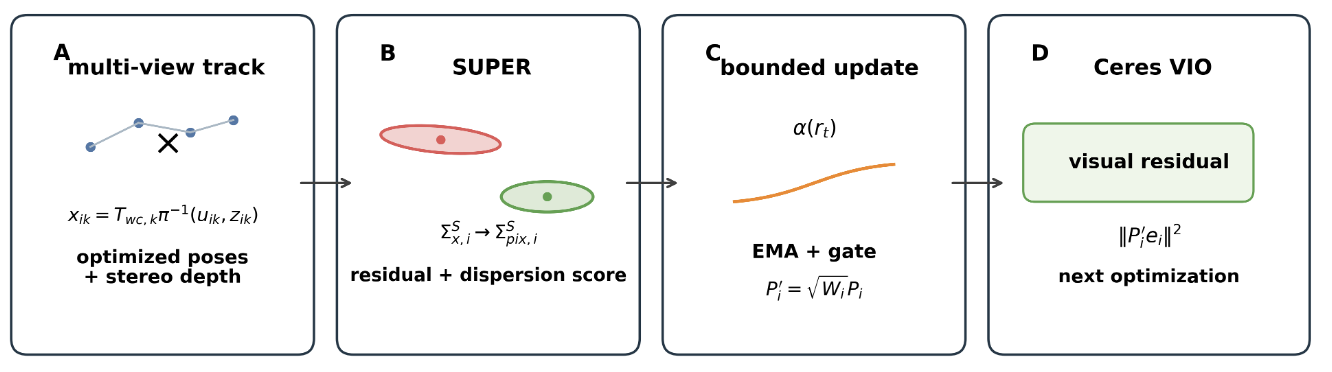}
    \caption{\textbf{Overview of SUPER:}
    SUPER forms a multi-view dispersion signal on track level on the basis of stereo observations and optimized backend states. These signals are mapped into the pixel space by considering the uncertainties. Next, the obtained information is combined with the residual to activate the temporal update for the ViO optimization in the subsequent timestep.}
    \label{fig:super_pipeline}
\end{figure}
As Fig.~\ref{fig:super_pipeline} illustrates, \method computes the reproejection residuals, landmark dispersion values from multi-view perspectives and the discrepancies from the projection geometry to produce a representation for the uncertainty on feature level. By complementing these steps with a protocol, \method additionally provides an activation gating, temporal smoothing within a confidence interval and an information floor with parameters frozen after calibration. We distinguish nominal operation, observations that are degraded but traceable, and the effect due to frontend collapse. Therefore, \method targets the conflicting regime, where persistent tracks remain valid but heterogeneous. Especially, the majority of existing robotics literature do not focusing on the derivation of measures to cope with effects from invalid associations, IMU faults or frontend collapse that cannot be handled by reweighting schemes in the backend optimizer. 
Our framework SUPER (Sensitivity-based Uncertainty-aware PErformance and Risk assessment) offers the possibility to assess the risk of visual odometry controllers at runtime. The performed experiments study the consequence of different visual disturbances and suggest metrics to overcome these. 
The remainder of this paper is structured as follows: First, Section \ref{sec:sota} summarizes reated works, thereby emphasizing the scientific contributions of this manuscript. Afterward, Section \ref{sec:method} introduces our framework SUPER in Section \ref{sec:method}. Section \ref{sec:experiments} presents the validation experiments. The respective reults are summarized and discussed in Section \ref{sec:results} before concluding this work in Section \ref{sec:discussion_conclusion}. 

\section{RELATED WORK}
\label{sec:sota}
\textbf{VO/VIO and Estimation via Learning}
Filter-based systems such as MonoSLAM~\cite{Davison2007} and MSCKF~\cite{Mourikis2007} propagate uncertainty explicitly, while optimization-based systems such as OKVIS~\cite{Leutenegger2015}, ROVIO~\cite{Bloesch2017}, and DROID-SLAM~\cite{Teed2021} use geometric, photometric, or learned representations. Learning-based approaches including D3VO~\cite{Yang2020}, DeepVO~\cite{Wang2017DeepVO}, and differentiable filters~\cite{Wagstaff2022} depend on trained models. 
While state estimators that use learning are computationally efficient, their shortcoming lies in the missing explainability. This, in turn, might result in the underestimation of risks in safety-critical situations. Therefore, \method derives its control signal online from the deployed optimizer.

\textbf{Robust and Adaptive Measurement Weighting.}
Robust kernels, switchable constraints~\cite{Suenderhauf2012}, dynamic covariance scaling (DCS)~\cite{Agarwal2013}, max-mixtures~\cite{Olson2013MaxMixturesIJRR}, and adaptive covariance inflation~\cite{Mehra1970AdaptiveKF} suggest methods to handle the undesired influence of inconsistent measurements. In particular, their control variables are formulated as residual magnitudes, latent switches, mixture components, or externally parameterized noise. By doing so, valuable information that originate from the exact type of the origin may be neglected. Therefore, \method maintains per-feature visual information from residual misfit and the directional disagreement of repeated triangulations. 

\textbf{Uncertainty in SLAM and Degeneracy Analysis}
Factor-graph tools such as GTSAM~\cite{Dellaert2012GTSAM}, iSAM2~\cite{Kaess2012iSAM2}, and g2o~\cite{Kuemmerle2011g2o} permit marginal-covariance recovery, while information and observability analyses characterize estimator conditioning~\cite{Barfoot2017SER,Thrun2005PR,Huang2010ObservabilitySLAM}. On the other hand, robustness benchmarks~\cite{Hendrycks2019ImageNetC} and learned uncertainty models~\cite{Kendall2017BayesSeg,Ilg2018FlowUncertainty} address complementary statistical questions. Our approach focuses on the empirical consistency among the optimized poses, stereo depths, and persistent feature tracks, thereby distinguishing from works that refer to marginal recovery or learned calibration; models. 

\textbf{Robust Weighting and Risk-awareness in Robotics}
Robust kernels, switchable constraints~\cite{Suenderhauf2012}, DCS~\cite{Agarwal2013}, max-mixtures~\cite{Olson2013MaxMixturesIJRR}, and adaptive filtering~\cite{Mehra1970AdaptiveKF} modify measurement influence by means of residuals, switches, mixtures, or noise parameters. Factor-graph and observability methods characterize the estimator uncertainty~\cite{Dellaert2012GTSAM,Kaess2012iSAM2,Kuemmerle2011g2o,Barfoot2017SER,Huang2010ObservabilitySLAM}, while risk-aware methods refer to uncertainty downstream~\cite{Blackmore2011}. 
In a broader sense, risk-awareness in robotics refers to the derivation of uncertainty or confidence bounds in planning, navigation, or control~\cite{Baek2023_ICRA, Gaus2025}. For example, Blackmore et al in 2011 develop a framework for chance-constrained planning in~\cite{Blackmore2011}. In contrast to these contributions that suggest to directly implement the uncertainty signal in the robot control, \method reflects the risk via an activation signal for the backend optimizer adaptation. 
In fact, \method considers the consistency between multiple viewpoints to control per-feature visual information, thereby represents the risk as an activation threshold.
The majority of above contributions is performed offline and refer to marginal statistics or held-out perturbation response. Instead of providing a signal for the optimization loop, \method closes the loop by adapting visual information during optimization. To consider undesired effects due to the nondeterministic character of the optimizer, we further propose an action-verification protocol that uses factor logs, inactive controls and linearization replay. Therefore, the scientific contributions of this paper can be summarized as follows: 
\begin{enumerate}
    \item We perform a sensitivity-aware uncertainty propagation that extracts per-feature pixel covariances and accumulates these into the feature level.
    \item By referring to the obtained uncertainty, we propose an action-verification protocol considering factor logs, inactive controls, and linearization replay to distinguish whether undesired variations in the trajectories originaete from the optimizer's nondeterminism.
\end{enumerate}
As the following sections will elaborate, \method provides a bounded VIO information controller by considering the residual misalignment and projected multi-view track dispersion. In this paper, \method is implements in Ceres and g2o. We perform evaluation experiments on a frozen policy with prospective replication and comparing different modes. 

\section{SUPER-- Sensitivity-based Uncertainty-aware Performance and Risk Assessment}
\label{sec:method}

%\begin{figure*}[t]
%  \centering
%  \includegraphics[width=0.92\textwidth]{RA-L2_img/260916_Fig2_pipeline.png}
 % \caption{\textbf{Schematic overview of the SUPER framework:}
%  The VIO backend provides optimized poses, persistent landmarks, stereo depths, and visual residuals.
%  SUPER converts the residual and the multi-view consistency quantities into a frame-level activation %signal and a dispersion proxy per feature and per pixel space. The closed loop updates a bounded
%  estimate for the state by referring to the measurement data for each landmark. Next, the obtained %information matrix is integrated into the visual residuals in the Ceres or g2o backend. The frozen Ceres %policy initialization is performed at $\rho_t\geq1$ and avoids the effect of overweighted information.}
 % \label{fig:method_overview}
%\end{figure*}

\begin{figure*}[t]
 \centering
 \includegraphics[width=0.9\textwidth]{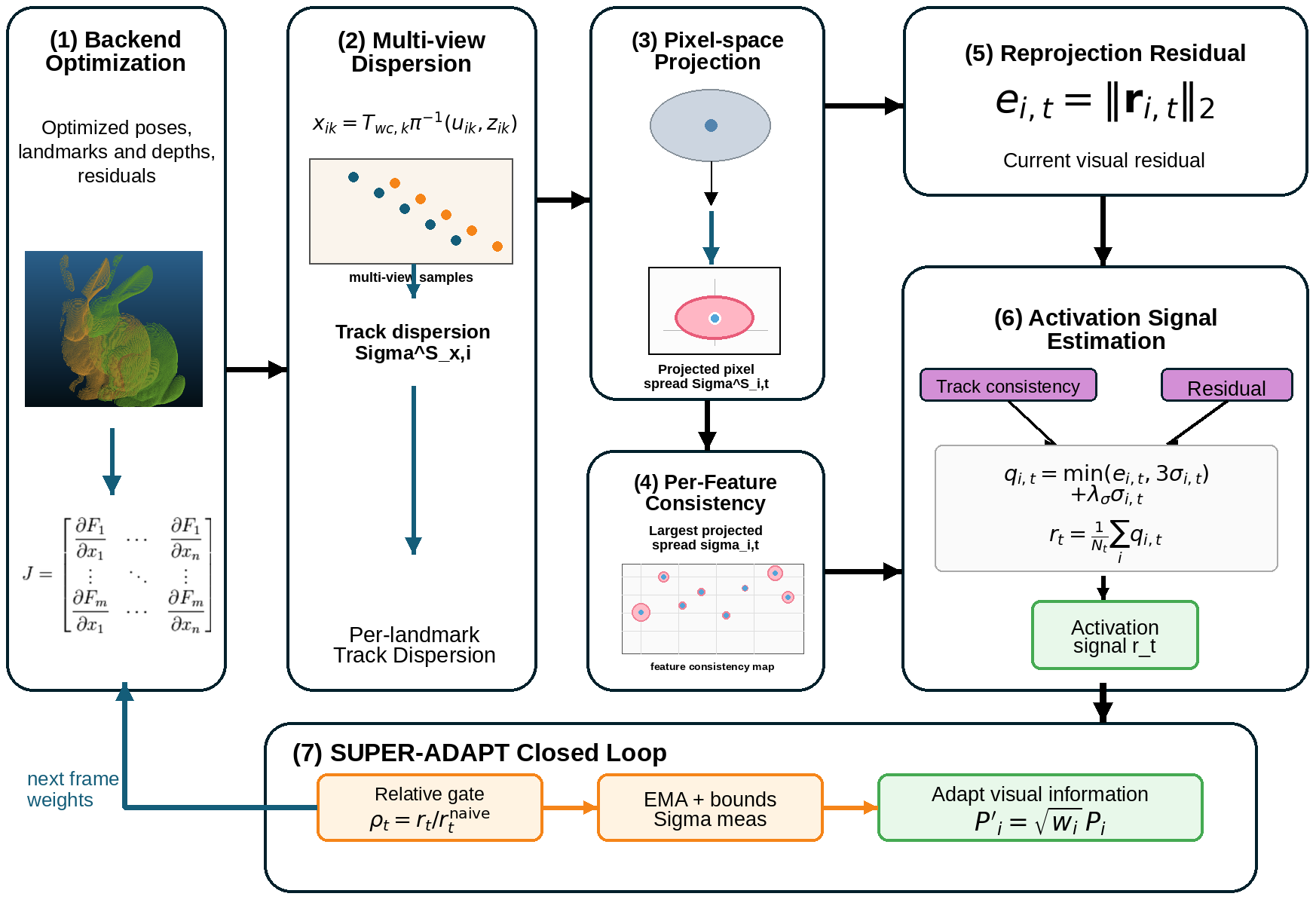}
\caption{ \textbf{Schematic overview of \method.}
\textbf{(1)} Backend optimization provides state estimates with residuals and Jacobians that form %the normal matrix.
\textbf{(2)} The Schur complement marginalizes poses and yields per landmark covariance.
\textbf{(3)} Each covariance is projected into the pixel space to obtain per feature uncertainty.
\textbf{(4)} Aggregating all pixel level covariances produces a dense map of feature uncertainties across the image.
\textbf{(5)} Geometric conditioning is computed from the shape of each pixel covariance, revealing how viewing geometry affects estimation quality.
\textbf{(6)} The residual and the feature uncertainty are transformed into an activation signal.
\textbf{(7)}the obtained information matrix is integrated into the visual residuals in the Ceres or g2o backend. The frozen Ceres policy initialization is performed at $\rho_t\geq1$ and avoids the effect of overweighted information.
}
\label{fig:method_overview}
\end{figure*}

\method computes a track-consistency proxy from stereo observations and optimized backend states, projects it into pixel space, and combines it with the reprojection residual to derive a bounded information update (Fig.~\ref{fig:method_overview}). The signal serves as the input for the controller input. %and therefore does not claim the recovery of marginal covariances.

\subsection{Track-Consistency Signal}
\label{sec:track_consistency}

Let landmark $i$ have $K_i$ valid stereo observations in the current sliding window. Observation $k$ provides an undistorted image location $u_{ik}$, stereo depth $z_{ik}$, and optimized camera pose $T_{wc,k}$. Back-projecting each observation into the world frame gives 
\begin{equation}
    x_{ik}
    =
    T_{wc,k}\,\pi^{-1}(u_{ik},z_{ik}),
    \label{eq:super_backprojection}
\end{equation}
where $\pi^{-1}(\cdot)$ denotes stereo back-projection.

For optimized landmark position $\hat{x}_i$, \method measures disagreement among these observation-wise reconstructions using
\begin{equation}
    \Sigma^{S}_{x,i}
    =
    \operatorname{diag}\!\left[
    \frac{1}{K_i-1}
    \sum_{k=1}^{K_i}
    (x_{ik,d}-\hat{x}_{i,d})^2
    \right]_{d\in\{x,y,z\}}
    + \epsilon I_3 .
    \label{eq:super_landmark_dispersion}
\end{equation}
This quantity is an empirical multi-view dispersion conditioned on the actual pose and stereo-depth estimates. At least three valid observations are required. For $K_i<3$, $\Sigma^{S}_{x,i}$ is set to the fixed fallback $\Sigma^{S}_{\mathrm{fb}}=\operatorname{diag}( \sigma^2_{\mathrm{fb},x}, \sigma^2_{\mathrm{fb},y}, \sigma^2_{\mathrm{fb},z})$. The same configured eigenvalue bounds are then applied as for valid multi-view dispersion estimates. For its actual observation, the world-frame dispersion is rotated into the camera frame and projected into pixel space:
\begin{equation}
    \Sigma^{S}_{i,t}
    =
    J_{\pi,i,t} R_{cw,t}
    \Sigma^{S}_{x,i}
    R_{cw,t}^{\top} J_{\pi,i,t}^{\top},
    \label{eq:super_pixel_covariance}
\end{equation}
where $R_{cw,t}$ rotates world-frame perturbations into the current camera frame and $J_{\pi,i,t}\in\mathbb{R}^{2\times3}$ is the reprojection Jacobian. Thus, $\Sigma^{S}_{i,t}\in\mathbb{R}^{2\times2}$ is the pixel-space track-consistency proxy used by the controller.

For activation, \method combines this geometric disagreement with the current reprojection residual. We define
\begin{equation}
    \sigma_{i,t}
    =
    \sqrt{
    \lambda_{\max}
    \left(
    \Sigma^{S}_{i,t}
    + \sigma_0^2 I_2
    \right)},
    \label{eq:super_sigma}
\end{equation}
and
\begin{equation}
    e_{i,t}
    =
    \|\mathbf{r}_{i,t}\|_2 ,
    \label{eq:super_residual}
\end{equation}
where $\sigma_0$ is the nominal pixel uncertainty and $\mathbf{r}_{i,t}$ is the visual reprojection residual. Each valid tracked observation contributes
\begin{equation}
    q_{i,t}
    =
    \min(e_{i,t},3\sigma_{i,t})
    +
    \lambda_\sigma \sigma_{i,t},
    \label{eq:super_feature_score}
\end{equation}
and the frame-level activation score is
\begin{equation}
    r_t
    =
    \frac{1}{N_t}
    \sum_{i=1}^{N_t}
    q_{i,t},
    \label{eq:super_raw_score}
\end{equation}
where $N_t$ is the number of valid tracked observations at frame $t$. Residual clipping limits the influence of a single inconsistent feature. The score $r_t$ is a dimensioned controller input, not a calibrated probability of estimator failure.

\subsection{Adaptive Visual Measurement Model}
\label{sec:adaptive_measurement_model}

The fixed baseline uses an isotropic visual information matrix
\begin{equation}
  \Lambda^{0}_{i} = \sigma_0^{-2} I_2 .
\end{equation}

We distinguish the instantaneous consistency proxy $\Sigma^{S}_{i,t}$ from the persistent controller state $\Sigma^{M}_{i,t}$. The former is recomputed from the current optimized track geometry, whereas the latter is temporally filtered and determines the information applied to subsequent visual factors.

\method sets the instantaneous target to
\begin{equation}
  \Sigma^{\star}_{i,t} = \beta\,\Sigma^{S}_{i,t},
  \label{eq:target_covariance}
\end{equation}
where $\beta$ is a global scale. The state is updated by an
exponential moving average:
\begin{equation}
  \widehat{\Sigma}^{M}_{i,t+1}
  =
  (1-\alpha_t)\Sigma^{M}_{i,t}
  +
  \alpha_t \Sigma^{\star}_{i,t}.
  \label{eq:ema_covariance}
\end{equation}
The adaptation speed depends on the actual frame-level activation given bt
\begin{equation}
  \alpha_t =
  \alpha_{\min}
  +
  (\alpha_{\max}-\alpha_{\min})
  \frac{1}{1+\exp(-\gamma r_t)} .
  \label{eq:risk_alpha}
\end{equation}
We use $\alpha_{\min}=0.01$, $\alpha_{\max}=0.30$, and $\gamma=1\,\mathrm{px}^{-1}$. At the observed active-frame scores, Eq.~\eqref{eq:risk_alpha} saturates at $\alpha_{\max}$ in the frozen Ceres policy, so the gate determines whether, not how rapidly, it intervenes; the sigmoid form is retained because the mechanism in Sec.~\ref{sec:ablation_results} varies $\alpha_{\max}$ over $[0.15,0.45]$ within the same control law. The activation gate equals
\begin{equation}
  (\Lambda_{i,t},\Sigma^{M}_{i,t+1}) =
  \begin{cases}
    (\Lambda^{0}_{i},\Sigma^{M}_{i,t}), &
    \rho_t < \rho_{\mathrm{gate}},\\
    (\operatorname{Info}(\widehat{\Sigma}^{M}_{i,t+1}),\widehat{\Sigma}^{M}_{i,t+1}), &
    \rho_t \ge \rho_{\mathrm{gate}}.
  \end{cases}
  \label{eq:risk_gate}
\end{equation}

To allow the comparison between datasets, the frozen policy refers to
\begin{equation}
  \rho_t =
  \frac{r_t}{r^{\mathrm{naive}}_t+\epsilon_\rho},
  \qquad
  \rho_t \ge \rho_{\mathrm{gate}},
  \label{eq:relative_gate}
\end{equation}
with $\rho_{\mathrm{gate}}=1.00$. The reference $r_t^{\mathrm{naive}}$ is computed from Eq.~\eqref{eq:super_raw_score} using the same residuals and tracked observations as $r_t$, but replacing the projected multi-view dispersion by the front-end's nominal pixel uncertainty. Hence $\rho_t$ measures the current consistency score relative to the score expected under the nominal measurement model, rather than against a dataset-specific absolute threshold. $\operatorname{Info}(\cdot)$ applies the bounds, inversion, and conditioning of Sec.~\ref{sec:bounds_conditioning_gate}. Below threshold, baseline information and the latent state are retained. Updates are causal: frame $t$ is optimized with $\Sigma^M_{i,t}$, and Eqs.~\eqref{eq:ema_covariance}--\eqref{eq:relative_gate} commit the state for frame $t+1$. %The information matrix is never recursively modified within the solve that supplies these states. 
The frozen Ceres policy uses $\sigma_0=1$\,px, $\beta=1$, $\lambda_\sigma=2$, $\rho_{\mathrm{gate}}=1$, $\epsilon=10^{-6}\,\mathrm{m}^2$, world-dispersion bounds $[10^{-5},10^{-1}]\,\mathrm{m}^2$ on the eigenvalues of $\Sigma^{S}_{x,i}$, covariance cap $\lambda_{\max}=5\,\mathrm{px}^2$ on $\Sigma^{M}_{i,t}$, information cap $10^6\,\mathrm{px}^{-2}$, $\eta=0.01$, $\kappa_{\mathrm{th}}=100$, and $w_{\min}=0.50$; it forbids information upweighting. The choice $\beta=1$ preserves the projected pixel-dispersion scale and is fixed before confirmation data are evaluated. Configuration values are type- and range-checked before execution, and invalid projected sources revert to fixed information. The separately frozen g2o transfer employs a temporally normalized gate of $1.10$, baseline-update rate $0.002$, an active EMA rate $0.30$, $\beta=1$, covariance cap $5$, and a $0.98$ baseline-information floor. 

\subsection{Bounds and Conditioning Gate}
\label{sec:bounds_conditioning_gate}

Prior to performing the inversion, the covariance is symmetrized and spectrally bounded according to
\begin{equation}
  \Sigma \leftarrow Q
  \operatorname{diag}\!\left(
  \operatorname{clip}(\lambda_j,\epsilon,\lambda_{\max})
  \right)Q^{\top}.
  \label{eq:eigen_clamp}
\end{equation}
The bounded covariance condition value is calculated via
\begin{equation}
  \tilde{\kappa}^{M}_i =
  \frac{\lambda_{\max}(\Sigma^{M}_{i,t+1})}
       {\lambda_{\min}(\Sigma^{M}_{i,t+1})}
  \label{eq:covariance_condition}
\end{equation}
and the controller applies a soft conditioning gate
\begin{equation}
  w_i =
  \max\!\left(
  w_{\min},
  \frac{1}{1+\eta\max(0,\tilde{\kappa}^{M}_i-\kappa_{\mathrm{th}})}
  \right).
  \label{eq:soft_gate}
\end{equation}
The final adaptive information matrix is computed as follows:
\begin{equation}
  \Lambda_{i,t}^{A}
  =
  w_i\,(\Sigma^{M}_{i,t+1})^{-1}.
  \label{eq:adaptive_information}
\end{equation}
The frozen no-upweight policy additionally clips every information eigenvalue to at most $\sigma_0^{-2}$. Together with the covariance cap, $w_{\min}$ prevents removal of all visual support.

\subsection{Insertion into Ceres and g2o}
\label{sec:backend_insertion}

In Ceres, the sliding-window reprojection factors take the square-root form of Eq.~\eqref{eq:adaptive_information}. Pose-only/PnP factors preserve baseline information in the implementation. \method refreshes the state after the sliding-window optimization. In ORB-SLAM3, the same track-dispersion construction is obtained from the map observations and considered in the information matrices of g2o mono and stereo visual edges. A conservative backend-specific duty guard limits the activation to two consecutive frames followed by a 60-frame cooldown. An information floor retains at least $98\%$ of the baseline. 

%The update is linear in the number of observations and operates on $2\times2$ matrices; Sec.~\ref{sec:runtime_diagnostics} reports its measured cost.

\section{Experimental Setup}
\label{sec:experiments}
This section aims to study following three hypotheses: 1) Hypothesis 1: The frozen adaptation limits nominal intervention., Hypothesis 2: Selective gains occur while visual support remains valid and Hypothesis 3: The combined consistency signal offers a distinct accuracy--stability trade-off from scalar controls. We refer to stereo Ceres \cite{Agarwal_Ceres_Solver_2022} as our primary platform and adapt the VIO pipeline from~\cite{Choi2025} with FAST features, stereo triangulation, IMU propagation, and sliding-window optimization. Secondly, ORB-SLAM3 stereo-inertial tracking~\cite{Campos2021ORB-SLAM3} supplies an auxiliary g2o backend with a distinct feature-association and optimization pipeline.

\subsection{Datasets and Corruptions}
\label{sec:noise_experiments}
The controlled discovery experiments are conducted on EuRoC MAV~\cite{Burri2016EuRoC} sequences MH\_01, MH\_03, MH\_05, and V1\_03 and TUM-VI~\cite{Schubert2018TUMVI} room1, room3, and room6. A prospective confirmation freezes the policy and corruption levels prior to evaluating the EuRoC sequences MH\_02, MH\_04, V1\_01, V1\_02, and V2\_01--03. Unmodified UMA-VI \cite{zuniga2020vi} and TUM-VI room2/4/5 serve to study natural recordings. TartanAir~\cite{tartanair2020iros} \emph{abandonedfactory} and \emph{gascola} are used to explore the execution on out-of-domain cases. The controlled matrix contains 17 predefined conditions: Clean input, spatial and temporal motion blur, JPEG compression, downsampling, temporal and moving rectangular occlusion, and combined blur/noise or blur/occlusion. The occluders isolate geometry-dependent partial track corruption and are considered as controlled stressors.

\subsection{Compared Modes}
\label{sec:compared_modes}

We primarily compare between \emph{fixed} and \emph{frozen} adaptive information under an identical frontend and robust loss. In the ablations, we remove controller components, vary bounds and covariance sources, and compare DCS-like, residual-only, and feature-count controls. 
To study the significance of the geometric signal, we additionally introduce a signal-blind random control matched to \method's duty rate. We refer to it as the \emph{duty-matched random-action control}. Every pair shares its input, corruption seed, frontend, and evaluator. The matched five-sequence study uses clean input, temporal blur, and temporal occlusion with three repetitions: 90 Ceres executions compare fixed/active, while 135 g2o executions add an inactive-instrumented control that preserves baseline information. Three additional suites test severity and diagnostic guards (471 executions), covariance/EMA/g2o-authority mechanisms (684), and exact Ceres replay, UMA-VI repeats, and natural g2o execution (186). Frozen policies and fixed seeds are retained as applicable; shadow branches modify only visual information while sharing state, factors, IMU terms, losses, constraints, and outlier masks. Visual information is the only manipulated variable in the principal Ceres comparison; thresholds, robust losses, and the frozen policies remain fixed outside named ablations. Policy selection minimizes clean-data interference, divergence, rotation damage, Ceres residual/Jacobian warnings, and runtime before considering translation gain. Parameters are selected only on MH05, V1\_03, TUM-VI room3 controlled cases and four UMA-VI clean sequences. The locked matrix, held-out and prospective sequences, repeats, mechanism controls, and natural/OOD data never influence policy choice.

\subsection{Metrics and Diagnostics}
\label{sec:metrics}
We report translational and rotational RMSE, with $\Delta=100(m_A-m_F)/m_F$. Negative values favor adaptation. The identical paired values reflect ties while translation RMSE above $1000$\,m is reported separately as a divergence. We aggregate the repetitions by sequence--condition cluster, with medians, W/L/T counts, positive-tail quantiles, and 95\% intervals from clusters that contain 10k bootstrap samples. We use paired sign and Wilcoxon tests, Benjamini--Hochberg correction within the nine corruption-family tests and conservatively across all 22 split/family tests, and Cochran's $Q$/$I^2$ for heterogeneity.  Factor-level logs and same-linearization replay are checked before trajectory changes are attributed to the controller.

\section{Results}
\label{sec:results}

\begin{figure*}[t]
  \centering
  \includegraphics[width=0.88\textwidth]{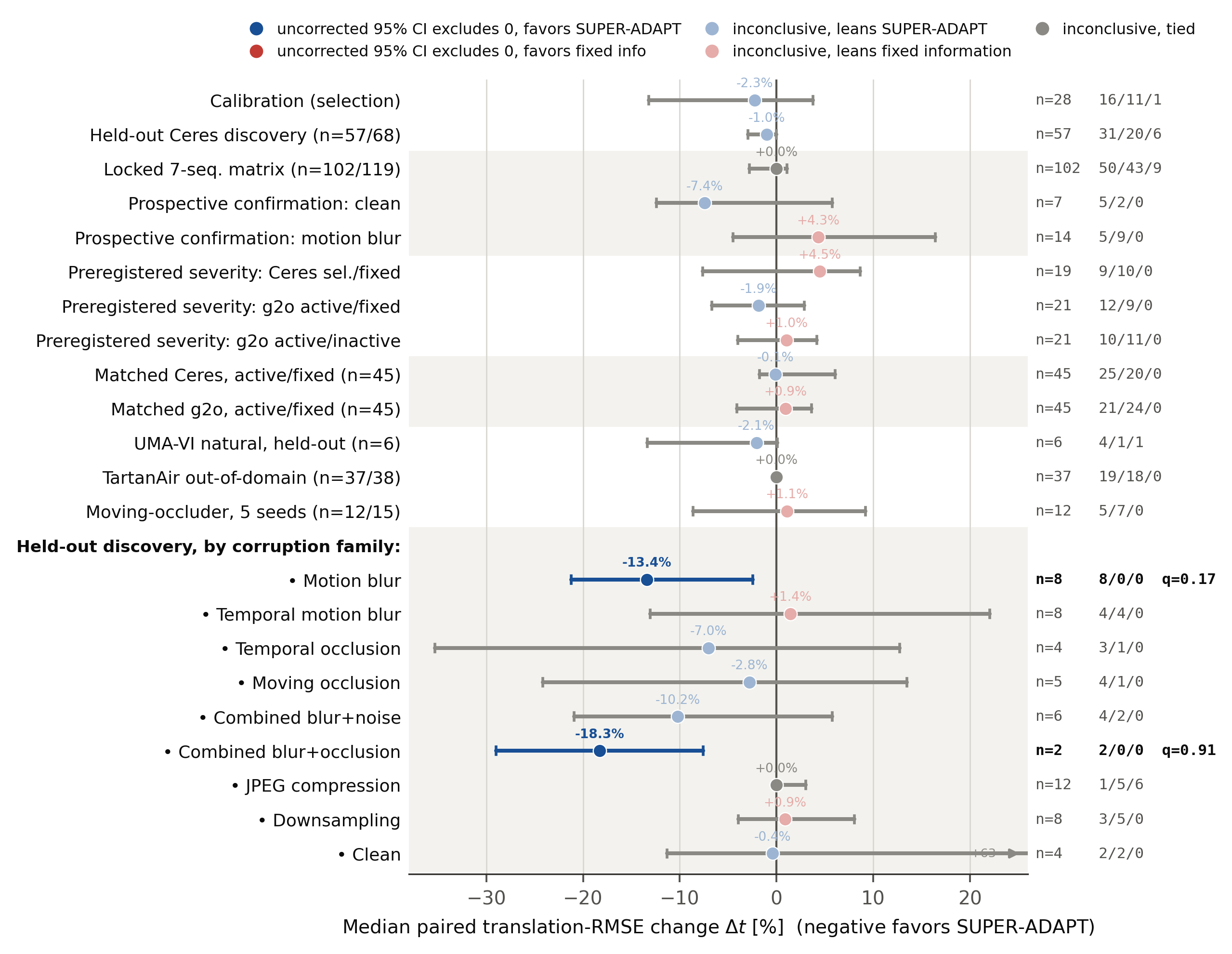}
  \caption{\textbf{Cross-split evaluation.} Median paired translation-RMSE change with 95\% cluster-bootstrap CIs; negative values favor \method. Top: 13 evaluation splits. Bottom: corruption-family decomposition of the held-out Ceres split.}
  \label{fig:forest_summary}
\end{figure*}

Fig.~\ref{fig:forest_summary} summarizes all evaluation splits. All 13 confidence intervals cross zero while the held-out corruption-family decomposition is heterogeneous. %Motion blur is the strongest nominal family result, but seems weak on FDR correction or prospective replication. Thus, no accuracy claim survives both correction and replication.

\subsection{Discovery and Prospective Ceres VIO}
\label{sec:main_vio_results}
%Calibration selects $\rho_{\mathrm{gate}}=1.00$ from seven candidates using solver stability, clean interference, rotation damage, and nondegenerate action duty prior to translational gain. On 28 calibration pairs, the selected policy provides 16/11/1 W/L/T, $-2.27\%$ median translation, and $-3.23\%$ median rotation change. The policy is then frozen. On the sequence-held-out discovery split, 57/68 pairs are robustly comparable and the remaining 11 diverge in both modes. \method yields 31/20/6 W/L/T, $-1.01\%$ median translational variation (95\% bootstrap CI $[-2.94,0.00]\%$), $-2.20\%$ mean translational variation, $-0.52\%$ median rotational variation, and reduces Ceres residual/Jacobian evaluation warnings from 99 to 60. Neither the sign test ($p=0.161$) nor a Wilcoxon signed-rank test on the paired percentage changes ($p=0.111$) rejects the null at this split size. Motion blur results in an exploratory 8/8 and a median variation of $-13.39\%$, while MH03 temporal occlusion 0.50 improves by $35.31\%$; Fig.~\ref{fig:shadow_replay_case} verifies this specific case with exact same-linearization replay.

Calibration selects $\rho_{\mathrm{gate}}=1.00$ from seven candidates using stability and clean-interference criteria before translation gain; the policy is then frozen. On the held-out discovery split, 57/68 pairs are robustly comparable and \method gives 31/20/6 W/L/T with $-1.01\%$ median translation change (95\% CI $[-2.94,0.00]\%$); neither the sign test ($p=0.161$) nor Wilcoxon test ($p=0.111$) rejects the null. Motion blur is the strongest exploratory subgroup ($-13.39\%$), while Fig.~\ref{fig:shadow_replay_case} verifies controller action on the MH03 temporal-occlusion case. The locked matrix is neutral overall (102/119 robust pairs, 50/43/9 W/L/T, $0.00\%$ median translation change). Prospective testing gives $-7.42\%$ median translation on clean input but $+4.31\%$ under motion blur, with blur 25 regressing by $+16.41\%$. Action duty does not provide a reliable fallback signal: discovery duty predicts loss worse than chance (AUC $0.29$), and a discovery-calibrated $76\%$ threshold detects only 2/9 prospective blur losses while missing the largest failure. Duty is therefore used to verify intervention, not to predict benefit. The held-out corruption-family decomposition is heterogeneous ($Q=20.8$, $p=0.008$). Motion blur is the strongest exploratory result ($-13.4\%$, $p=0.008$), but fails family-level FDR correction ($q=0.07$) and regresses prospectively ($+4.31\%$).  Combined blur/occlusion is based on only two robust pairs and is likewise not significant after correction. We therefore claim heterogeneity across corruption families, not a family-level accuracy gain.

\begin{table}[t]
\centering
\caption{Frozen-policy confirmation. $R/P$ denotes robust/planned pairs;
the final column reports fixed/adaptive divergences and the Ceres-warning
delta $\Delta C$.}
\label{tab:split_confirmation}
\scriptsize
\setlength{\tabcolsep}{2.2pt}
\begin{tabular}{@{}lrrrr@{}}
\toprule
Split & $R/P$ & W/L/T & Med. $\Delta t$ & Div. F/A / $\Delta C$ \\
\midrule
Held-out Ceres & 57/68 & 31/20/6 & $-1.01\%$ & 11/11 / $-39$ \\
Locked Ceres & 102/119 & 50/43/9 & $0.00\%$ & 17/17 / $-63$ \\
Prospective clean & 7/7 & 5/2/0 & $-7.42\%$ & 0/0 / $-3$ \\
Prospective blur & 14/14 & 5/9/0 & $+4.31\%$ & 0/0 / $+5$ \\
Repeated all UMA-VI & 10/10 & 7/2/1 & $-2.07\%$ & 0/0 / -- \\
Repeated held-out UMA & 6/6 & 4/1/1 & $-2.07\%$ & 0/0 / -- \\
\bottomrule
\end{tabular}
\end{table}

\subsection{Component Ablations and Action Verification}
\label{sec:ablation_results}
Factor-level logs verify intervention throughout the frozen evaluation. On held-out clean data, median gate-open and modified-factor rates are $0.67\%$ and $0.59\%$; under degradation they rise to $17.69\%$ and $7.06\%$. Median invalid-source incidence is $0.12\%$ of calls, with invalid sources rejected before insertion. Thus, the controller is measurably and nontrivially active in these comparisons; causal attribution of local state increments is examined separately by same-linearization replay.

\begin{table}[t]
\centering
\caption{Ablations relative to fixed information. $R$ denotes robust pairs.}
\label{tab:simple_signal_baselines}
\scriptsize
\setlength{\tabcolsep}{2.0pt}
\begin{tabular}{@{}lrrrr@{}}
\toprule
Mode & $R$ & W/L/T & Med. $\Delta t$ & P90 $\Delta t$ \\
\midrule
\multicolumn{5}{c}{Scalar controls} \\
\method & 23 & 13/7/3 & $-2.14\%$ & $+11.64\%$ \\
Residual-only & 23 & 14/6/3 & $-2.03\%$ & $+14.70\%$ \\
Feature count & 23 & 12/8/3 & $-0.76\%$ & $+8.01\%$ \\
DCS-like & 23 & 8/9/6 & $0.00\%$ & $+100.75\%$ \\
Random-action (matched duty) & 23 & 9/11/3 & $+0.00\%$ & $+68.11\%$ \\
\midrule
\multicolumn{5}{c}{Controller components} \\
\method & 12 & 6/5/1 & $-3.82\%$ & $+20.23\%$ \\
Aggressive cap/floor & 12 & 9/2/1 & $-3.38\%$ & $+13.47\%$ \\
Conservative cap/floor & 12 & 8/3/1 & $-3.50\%$ & $+17.41\%$ \\
No gate & 12 & 7/4/1 & $-10.49\%$ & $+6.36\%$ \\
No EMA & 12 & 5/6/1 & $+0.19\%$ & $+11.63\%$ \\
Isotropic & 12 & 3/8/1 & $+7.17\%$ & $+92.67\%$ \\
Raw target & 12 & 7/4/1 & $-3.68\%$ & $+33.18\%$ \\
\midrule
\multicolumn{5}{c}{Scale and upweight controls} \\
$\beta=0.5$ & 11 & 6/4/1 & $-4.13\%$ & $+7.88\%$ \\
$\beta=1$ & 11 & 8/2/1 & $-8.05\%$ & $+2.56\%$ \\
$\beta=2$ & 11 & 7/3/1 & $-0.77\%$ & $+1.45\%$ \\
Allow upweight & 11 & 7/3/1 & $-5.89\%$ & $+9.17\%$ \\
\midrule
\multicolumn{5}{c}{Covariance sources} \\
Dispersion & 7 & 5/1/1 & $-9.54\%$ & $+8.10\%$ \\
Local Hessian & 6 & 1/4/1 & $+5.30\%$ & $+28.72\%$ \\
Ceres marginal & 6 & 3/2/1 & $-1.72\%$ & $+26.96\%$ \\
\bottomrule
\end{tabular}
\end{table}

No scalar baseline dominates \method (Table~\ref{tab:simple_signal_baselines}). The duty-matched random-action control opens its gate at \method's median measured duty on this set ($6.1\%$), independently of the \method signal, and applies a uniform random isotropic scale in $[1,5]\,\mathrm{px}^2$. This control tests whether bounded downweighting at the same intervention rate can reproduce \method's effect without using the geometric signal. Residual-only has a similar median but larger tail, while DCS-like and duty-matched random intervention show substantially heavier positive tails. The random control has no median benefit and a P90 loss tail about six times \method's, although their direct paired contrast is not significant ($p=0.55$). Removing the gate increases solver warnings, and the larger parameter sweeps do not identify a universally favorable setting.

\begin{figure}[t]
  \centering
  \includegraphics[width=\linewidth]{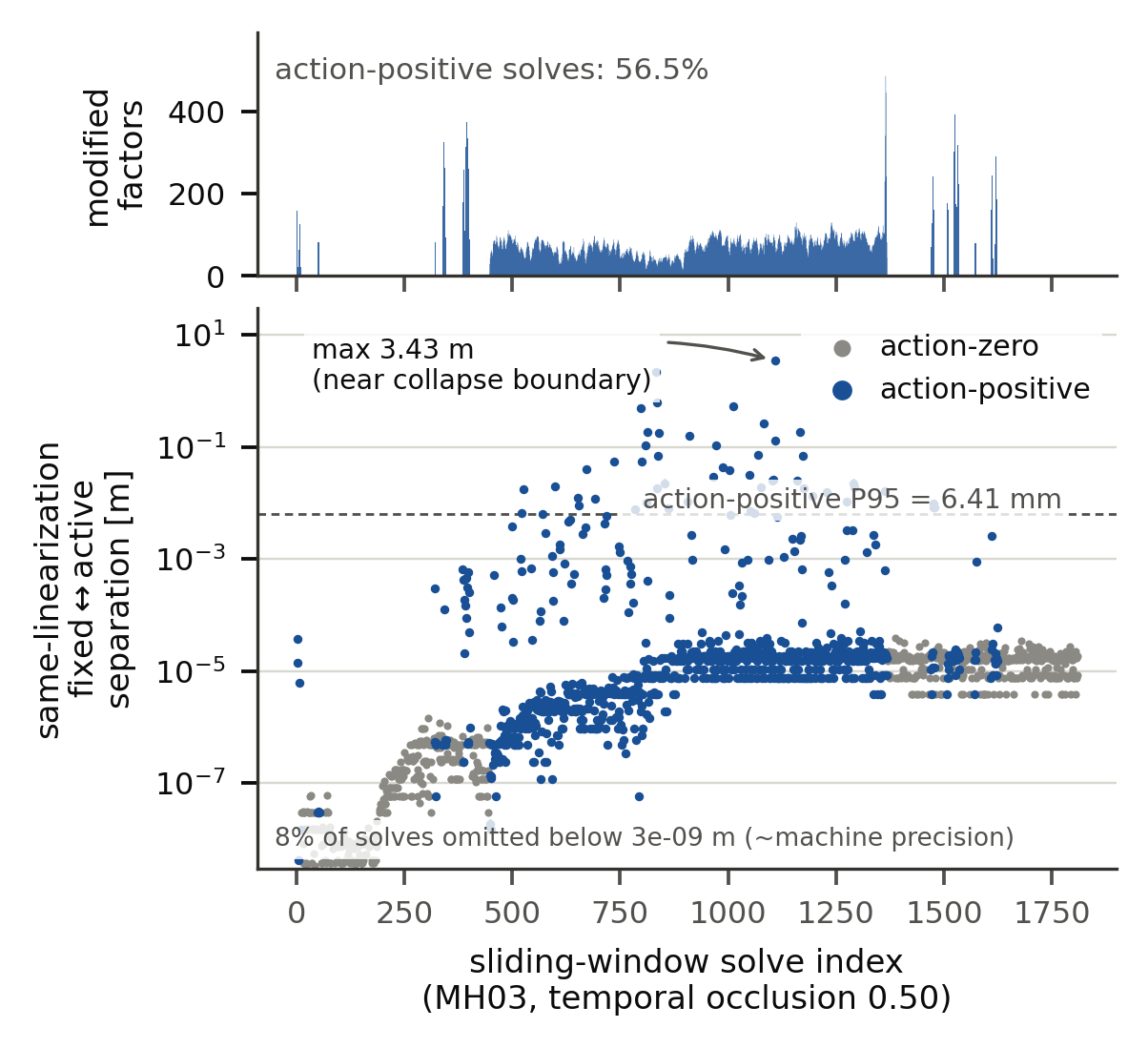}
  \caption{\textbf{Same-linearization replay, MH03 temporal occlusion and repeat.} Top: Modified visual factors per sliding-window solve. Bottom: The  fixed$\leftrightarrow$active pose separation from an identical shadow solve. The color indicates modifications. The action-zero solves correspond to floor precision while action-positive solves range from sub-millimeter to the $3.43$\,m outlier.}
  \label{fig:shadow_replay_case}
\end{figure}

Five identical-input repeats separate repeatable local effects from single-run outcomes. MH01 motion blur 15 improves in all five runs (median $-48.12\%$ translation, $-15.47\%$ rotation), and MH03 temporal occlusion 0.50 likewise improves in all five (median $-35.31\%$, $-25.00\%$). Conversely, MH01 clean worsens in all five (median $+23.14\%$ translation, $+252.97\%$ rotation). These repeats establish sequence--condition stability, not statistical independence. Replaying Ceres verifies the solver action. We distinguish the per-solve action rate and modified-factor duty. Across three repeats, the per-solve action rate spans $1.31$--$84.74\%$ over the four audited cases with action-conditioned 95th-percentile translation separations of $0.12$--$6.41$\,mm. Action-zero replay remains at numerical precision ($\approx2\times10^{-15}$\,m) while one MH03 solve reaches $3.43$\,m near collapse. Fig.~\ref{fig:shadow_replay_case} shows the effect for MH03 temporal occlusion 0.50: Modified-factor count and same-linearization fixed/active separation are plotted for every sliding-window solve, making the occasional large tail visible. The protocol does not depend on the controller. The identical same-linearization shadow-replay pipeline is repeatedly run on the MH03 temporal occlusion 0.50 with \method replaced by the residual-only scalar control (Sec.~\ref{sec:compared_modes}). Where \method acts selectively (duty $9.7\%$ on this case) and improves translation RMSE by $35.31\%$, residual-only modifies visual factors on every solve by construction and improves the same case ($17.71\%$). The verification correctly distinguishes \method's sparse, bounded intervention from residual-only's dense, unconditional one using the same factor logs. This demonstrates that the protocol generalizes across controllers. The larger preregistered mechanism suite qualifies the compact ablations. Over 14/15 robust covariance clusters, world-full, world-diagonal, pixel-diagonal, and isotropic modes provide 7/7, 5/9, 6/8, and 6/8 W/L with respective median changes of $+0.30\%$, $+2.39\%$, $+0.89\%$, and $+2.13\%$. Thus, neither the evaluated diagonal source nor anisotropy is uniquely favorable. Across 8/9 robust score/EMA clusters, $\lambda_\sigma\in[0,4]$ leads to overlapping intervals. 

\subsection{Matched Ceres/g2o Confirmation}
\label{sec:orb_results}

The matched study completes all 225 executions. Across 45 comparisons per backend, Ceres is near neutral (25/20 W/L, $-0.10\%$ median translation), but a $+397.14\%$ TUM-VI outlier raises its temporal-occlusion mean to $+62.06\%$; a neutral median therefore does not imply bounded tail risk. The g2o policy modifies edges in 25/45 runs at $0.013\%$ median duty over all runs ($0.014\%$ on the degraded subset, Table~\ref{tab:intervention_audit}). Table~\ref{tab:matched_backends} shows improvement under blur but regression under occlusion against both fixed and inactive controls; aggregate active/fixed performance is 21/24 and $+0.94\%$. The A/I clean contrast is omitted because the g2o policy is action-zero on clean input; active and inactive branches are therefore information-identical. Logged action supports mechanism transfer, not backend-independent accuracy improvement.
\begin{table}[t]
\centering
\caption{Matched three-repeat confirmation. A/F and A/I denote active/fixed and active/inactive comparisons.}
\label{tab:matched_backends}
\scriptsize
\setlength{\tabcolsep}{2.3pt}
\begin{tabular}{@{}llrrr@{}}
\toprule
Backend/contrast & Condition & W/L/T & Med. $\Delta t$ & Med. $\Delta R$ \\
\midrule
Ceres A/F & Clean & 9/6/0 & $-1.74\%$ & $+1.17\%$ \\
 & Temporal blur & 6/9/0 & $+6.05\%$ & $+1.13\%$ \\
 & Temporal occl. & 10/5/0 & $-0.10\%$ & $-3.50\%$ \\
\midrule
g2o A/F & Clean & 6/9/0 & $+2.29\%$ & $-0.26\%$ \\
 & Temporal blur & 10/5/0 & $-4.07\%$ & $+0.85\%$ \\
 & Temporal occl. & 5/10/0 & $+3.78\%$ & $+0.02\%$ \\
\midrule
g2o A/I & Temporal blur & 9/6/0 & $-10.77\%$ & $+0.15\%$ \\
 & Temporal occl. & 2/13/0 & $+3.64\%$ & $-0.12\%$ \\
\bottomrule
\end{tabular}
\end{table}
The causal/safety suite confirms no broad advantage (Table~\ref{tab:causal_safety}). Ceres is unfavorable in median; g2o is mildly favorable against fixed information but unfavorable against the inactive control. All three principal intervals cross zero. Action stratification prevents causal over-attribution: active/fixed action-positive runs give 20/21 W/L and $+0.54\%$, while action-zero runs give 10/9 and $-1.18\%$. Exact replay confirms a backend effect, but the selected policy modifies only 438 of 165,543 solves, with a median run-level 95th-percentile increment of $1.05\times10^{-6}$\,m. Its floor and duty guard therefore exert less influence on outcomes than frontend and initialization variability do. Authority is non-monotonic: floors $0.75$, $0.90$, $0.95$, and $0.98$ give medians of $+2.58\%$, $+2.36\%$, $-4.42\%$, and $+0.14\%$ against fixed. The nominally favorable $0.95$ floor remains indistinguishable from inactive control, and action-positive/zero subsets are similarly favorable. Replay increments stay at $1.20$--$1.41\times10^{-6}$\,m, precluding causal retuning from this isolated result. Diagnostic fallback is local rather than generally protective: the strict guard improves both MH04 clusters, but mild/strict guards worsen clean MH01 by $20.72\%$/$213.02\%$. All four TUM-VI occlusion clusters collapse in both modes and are excluded from accuracy claims; no guard preserves clean behavior while rescuing collapse.

\begin{table}[t]
\centering
\caption{Preregistered causal/safety suite. $R/P$ denotes robust/planned
sequence--condition clusters after repeat aggregation. Negative translation
change favors the first mode. TUM guard rows include clean clusters only;
two-cluster guard intervals are descriptive.}
\label{tab:causal_safety}
\scriptsize
\setlength{\tabcolsep}{1.0pt}
\begin{tabular}{@{}llrrr@{}}
\toprule
Backend/contrast & Scope & $R/P$ & W/L/T & Med. $\Delta t$ [95\% CI] \\
\midrule
Ceres selected/fixed & Severity & 19/21 & 9/10/0 & $+4.48$ [$-7.64,+8.67$] \\
g2o active/fixed & Severity & 21/21 & 12/9/0 & $-1.86$ [$-6.70,+2.89$] \\
g2o active/inactive & Severity & 21/21 & 10/11/0 & $+1.04$ [$-4.01,+4.16$] \\
Ceres strict/fixed & MH04 & 2/2 & 2/0/0 & $-11.13$ [$-13.67,-8.59$] \\
Ceres mild/fixed & TUM clean & 2/2 & 2/0/0 & $-6.12$ [$-11.97,-0.28$] \\
\bottomrule
\end{tabular}
\end{table}

\iffalse % Superseded inactive-branch diagnostic retained in source.
\begin{figure}[t]
  \centering
  \includegraphics[width=\linewidth]{RA-L2_img/fig_orb_status_by_sequence.png}
  \caption{ORB-SLAM3/g2o negative-control status accounting. Despite equivalent
  edge information, stochastic tracking and initialization produce one-sided
  successes and failures; these runs are not evidence for adaptation.}
  \label{fig:orb_status}
\end{figure}
\fi

\subsection{Performance OOD, Natural, Stress Scenarios}
\label{sec:ood_dynamic_results}

Evaluating all ten unmodified UMA-VI sequences results in 7/2/1 W/L/T and $-2.07\%$ median translational change. The six held-out sequences provide 4/1/1, $-2.07\%$ median translation and $-0.03\%$ median rotational change. Conference-csc3 and third-floor-csc2 improve by $14.15\%$ and $12.52\%$. Leave-one-sequence-out medians remain favorable ($-3.02$ to $-1.13\%$), but only five outcomes are non-tied. The calibrated elevator sequence improves the translation by $69.78\%$ but impedes the rotation by $18.33\%$.  Natural TUM-VI room2/4/5 results in 2/1/0 and $-7.41\%$ median change. Across 38 uncorrupted TartanAir pairs, 37 are robustly comparable and yield 19/18/0, $-0.02\%$ median and $+1.37\%$ mean translational variation. Easy Gascola improves by $30.91\%$ while hard trajectories regress in median. Moving-occluder outcomes remain geometry dependent. Across five seeds on MH05, V1\_03, and TUM-VI room3, 12/15 pairs are robust and provide 5/7/0, $+1.09\%$ median with $-3.48\%$ mean translational change. 

\subsection{Runtime and Failure Diagnostics}
\label{sec:runtime_diagnostics}
The dedicated subset results in $-0.014\%$ median wall-time and $1.33\%$ median per-frame overhead where available. Final-suite paired wall times remain within $\pm0.03\%$. The 471- and 684-execution suites complete 467 and 681 runs, respectively; seven g2o failures are retained in status accounting, and every sequence--condition cluster retains completed repeats. All 72 Ceres add-on executions succeed. After filtering one unmatched MH04 stereo timestamp, controlled g2o completes 24/24; its selected-policy MH04 runs remain action-zero and replay separation stays at numerical precision. Natural UMA-VI g2o is not comparable with regard to the accuracy: 17/90 executions terminate with signal 11, while the rest  align in 2--34 of roughly 900--1000 frames after IMU-initialization and map-reset failures occur. 
\begin{table}[t]
\centering
\caption{Median intervention duty from factor-level logs.}
\label{tab:intervention_audit}
\scriptsize
\setlength{\tabcolsep}{3.5pt}
\begin{tabular}{@{}lrrr@{}}
\toprule
Condition & Runs & Gate open & Modified factors \\
\midrule
Held-out clean & 4 & $0.67\%$ & $0.59\%$ \\
Held-out degraded & 64 & $17.69\%$ & $7.06\%$ \\
Matched Ceres clean & 15 & $0.53\%$ & $0.47\%$ \\
Matched Ceres degraded & 30 & $17.41\%$ & $15.57\%$ \\
Matched g2o clean & 15 & $0.00\%$ & $0.00\%$ \\
Matched g2o degraded & 30 & $0.069\%$ & $0.014\%$ \\
Repeated natural UMA-VI & 30 & $0.46\%$ & $0.46\%$ \\
TartanAir & 38 & $46.41\%$ & $23.19\%$ \\
\bottomrule
\end{tabular}
\end{table}

%The separation between clean and degraded duty verifies selective action, while high TartanAir duty without aggregate improvement shows that activation alone is not evidence of benefit.

\section{Discussion, Limitations, and Conclusion}
\label{sec:discussion_conclusion} 
We conclude that \method does not establish a general accuracy improvement. In all 13 split-level intervals cross zero, the factor-level logs and same-linearization replay show that the controller is  active. Repeat experiments further show that both gains and regressions can be stable on specific inputs. Also, alternative options do not provide more promising trade-offs. Limitations include the fact that the multi-view dispersion is conditioned on estimated poses and depths. In fact, our work focuses on the visual information of the sliding window. Furthermore, the validation experiments cover active Ceres and controlled g2o. In future, we plan to apply our framework on further additional estimator classes. In summary, \method is computationally efficient, bounded, and selectively active. While we note the dependency on the input, the action-verification protocol enables to distinguish the origins of the  resulting trajectory for backend-adaptive estimators that run inside nondeterministic optimizers. 

\bibliographystyle{IEEEtran}
\bibliography{references}
\end{document}